%% file: main.tex
\documentclass[runningheads]{llncs}

\usepackage{eccv}

\usepackage{eccvabbrv}

\usepackage{graphicx}
\usepackage{booktabs}
\usepackage{multirow}     

\usepackage[accsupp]{axessibility}  

\input{math_commands}

\usepackage{hyperref}

\usepackage{orcidlink}

\begin{document}

\title{Improving Hyperbolic Representations via Gromov-Wasserstein Regularization\thanks{Accepted by ECCV 2024.}}

\titlerunning{Improving Hyperbolic Reps via GW Reg}

\author{Yifei Yang\inst{1}\orcidlink{0009-0003-0529-9515} \and
Wonjun Lee\inst{2}\orcidlink{0000-0003-1280-7533} \and
Dongmian Zou\inst{3}\orcidlink{0000-0002-5618-5791} \and
Gilad Lerman\inst{2}\orcidlink{0000-0003-4624-3115}
}

\authorrunning{Y.~Yang \etal}

\institute{Wuhan University, Wuhan, Hubei 430072, China\\ \email{yfyang@whu.edu.cn}\\ 
\and
University of Minnesota, Minneapolis, MN 55455, USA\\
\email{\{lee01273,lerman\}@umn.edu}\\
\and
Duke Kunshan University, Kunshan, Jiangsu 215316, China\\ \email{dongmian.zou@duke.edu}
}
\maketitle


\begin{abstract}
Hyperbolic representations have shown remarkable efficacy in modeling inherent hierarchies and complexities within data structures. Hyperbolic neural networks have been commonly applied for learning such representations from data, but they often fall short in preserving the geometric structures of the original feature spaces. In response to this challenge, our work applies the Gromov-Wasserstein (GW) distance as a novel regularization mechanism within hyperbolic neural networks. The GW distance quantifies how well the original data structure is maintained after embedding the data in a hyperbolic space. Specifically, we explicitly treat the layers of the hyperbolic neural networks as a transport map and calculate the GW distance accordingly. We validate that the GW distance computed based on a training set well approximates the GW distance of the underlying data distribution. Our approach demonstrates consistent enhancements over current state-of-the-art methods across various tasks, including few-shot image classification, as well as semi-supervised graph link prediction and node classification.
\keywords{Gromov-Wasserstein distance \and Hyperbolic neural networks \and Few-shot learning \and Semi-supervised learning}
\end{abstract}

\section{Introduction}\label{sec:intro}
The success of many machine learning applications relies on effectively capturing the geometric intricacies of complex data structures. This challenge necessitates a careful selection of the underlying space that aligns with the inherent properties of data. In this context, hyperbolic spaces have emerged as a popular choice for modeling data with hierarchical structures such as lexical databases~\cite{nickel2017poincare} and phylogenetic trees~\cite{sala2018representation}. This is attributed to their distinctive capacity to embody exponential growth and branching patterns, a feature rooted in their constant negative curvature. Such a geometric framework can reflect the tree-like architectures that are ubiquitous across numerous real-world networks. 

\begin{figure}[!t]
  \centering
  \includegraphics[width=\linewidth]{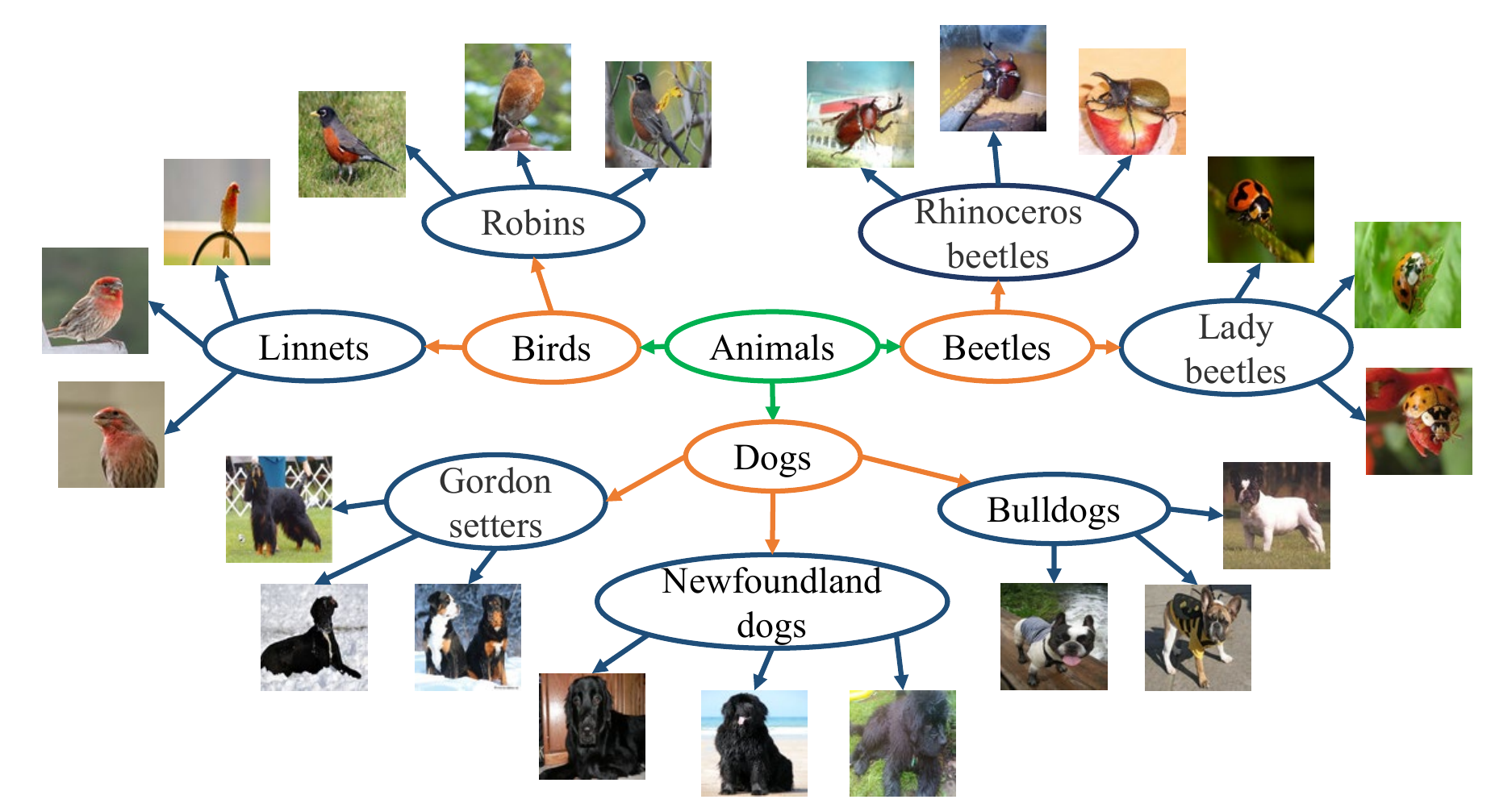}
  \caption{Illustration of hierarchical structures in image data. The images are taken from the MiniImagenet dataset.}
  \label{fig:image_data}
\end{figure}

Interestingly, hyperbolic spaces also play a significant role in computer vision~\cite{khrulkov2020hyperbolic, atigh2022hyperbolic}, particularly in tasks like image classification and object recognition where hierarchical relationships among instances and concepts are common. For example, within the broader category of animals, there are dogs, which further subdivide into breeds such as bulldogs, Gordon setters, Newfoundland dogs, \etc. This is illustrated in Fig.~\ref{fig:image_data}. Hyperbolic embeddings can capture these hierarchies, thereby potentially improving classification performance and providing more interpretable models.

In crafting hyperbolic representations from non-hyperbolic features, two main strategies emerge: principled design and data-driven learning. The principled approach, exemplified by methods like Sarkar's algorithm~\cite{sarkar2011low}, focuses on constructing embeddings that rigorously preserve the inherent geometric structure from data. This method particularly excels at maintaining hierarchical relationships within the data, ensuring that the embedded representation reflects structure of the original dataset with low distortion. On the other hand, the data-driven approach leverages the capabilities of end-to-end neural networks to learn the hyperbolic embedding directly from data. Such neural networks are well-known as hyperbolic neural networks (HNNs), which are pioneered in~\cite{ganea2018hyperbolic}. HNNs prioritize the development of hyperbolically coherent neural operations that are expressive and mirror the Euclidean counterparts. When dealing with input that is primarily Euclidean, HNNs employ straightforward operations, such as exponential maps, to transition features into the hyperbolic space, thereafter learning to extract deep features in a manner driven by the data. This method adapts the embedding to specific tasks, potentially unveiling intricate patterns and relationships inherent within data.

In our work, we aim to establish a methodology that integrates the benefits of low-distortion embedding into data-driven learning. In the context of learning where generalization is the goal, we can view hyperbolic embedding as a method of embedding the data distribution into hyperbolic spaces. Therefore, we need to compare the distributions before and after the embedding. This necessitates a comparison of data distributions before and after embedding, which is complicated by the fundamental differences in geometric principles, such as curvature and dimensionality, that govern Euclidean and hyperbolic spaces. Direct comparisons between such disparate spaces are inherently complex and challenging.

To address this challenge, we rely on the Gromov-Wasserstein (GW) distance \cite{memoli2007use}, a tool proven effective in comparing distributions across diverse metric spaces. Specifically, we employ the Gromov-Monge (GM) formulation, an alternative form of the GW distance, to compute the explicit transport map between heterogeneous metric spaces. The HNN layers induce the embedding distribution, thus serving as the transport map. Consequently, the GM formulation is a more suitable approach for the corresponding HNN layers. However, implementing the GM distance poses computational challenges due to its complex constraints. Therefore, in light of this, we extend the GM-based embedding technique introduced in~\cite{lee2023monotone} to compute the geometry-preserving transport map from data distribution in Euclidean space to hyperbolic spaces.

The contribution of the current work is summarized as follows:
\begin{enumerate}
    \item We propose a novel methodology that incorporates the GM distance into an end-to-end learning framework for hyperbolic representation learning. This approach ensures a more faithful representation of the intrinsic hierarchical structures. Moreover, we verify that our approach maintains computational efficiency by analyzing its time complexity.
    \item We validate that the embedding faithfully represents an embedding of distributions from the original domain by providing an upper bound for the sampling error. Therefore, regularization on the training sample generalizes to the underlying data distribution.
    \item We provide empirical evidence showing that our regularization strategy leads to consistent improvements over existing baseline methods when applied to datasets with hierarchical structures, including both image and graph data. 
\end{enumerate}

\section{Related Works}\label{sec:Related works}
\subsubsection{Hyperbolic embedding}
The adoption of hyperbolic geometries for representing hierarchical connections between data points has attracted considerable attention. The exploration began when researchers considered low-distortion embedding of tree or tree-like graphs into the hyperbolic domain. In their seminal work, Sarkar~\cite{sarkar2011low} showed that a tree can be embedded into the Poincar\'{e} disk with arbitrarily low distortion. Sala \etal~\cite{sala2018representation} generalized the embedding to high dimensional Poincar\'{e} balls and discussed embedding general graphs into trees. Sonthalia and Gilbert~\cite{sonthalia2020tree} bypassed the intermediary step of general graph structures, opting instead to directly infer tree architectures from the data itself. In addition to combinatorial methods, gradient-based methods are also used in hyperbolic embedding \cite{nickel2017poincare, nickel2018learning}.

\subsubsection{Hyperbolic neural networks}
More recently, the advancement in hyperbolic representation learning has evolved to be more closely aligned with specific end tasks and specific datasets, such as text \cite{tifrea2018poincar}, images \cite{khrulkov2020hyperbolic}, biology \cite{zhou2021hyperbolic}, molecular\cite{liu2019hyperbolic}, and knowledge graph\cite{chen2021fully}. This has led to the emergence of hyperbolic neural networks (HNNs) as a preferred framework for learning their representations. Numerous HNN architectures have emerged \cite{ganea2018hyperbolic, gulcehre2018hyperbolic, nickel2018learning, liu2019hyperbolic, chami2019hyperbolic, bachmann2020constant, shimizu2021hyperbolic}. Comprehensive surveys on HNN include~\cite{peng2021hyperbolic, yang2022hyperbolic}. These HNNs define operations that conform with a choice of coordinate system in the hyperbolic space. However, when communicating between the Euclidean features and their hyperbolic embeddings, these HNNs take very simple approaches. For instance, Ganea \etal~\cite{ganea2018hyperbolic} directly used logarithmic and exponential maps at the origin, leading to unavoidable distortions. There are also very recent works addressing representations learned by HNNs. For instance, CO-SNE~\cite{guo2022co} reduces the dimensions of hyperbolic features for visualization, which is an analogy to t-SNE in the Euclidean domain. Parallel to this, Fan \etal~\cite{fan2022nested} developed a systematic method for embedding data into nested hyperbolic spaces, which not only results in dimensionality reduction but also extends to design of HNNs. Nikolentzos \etal~\cite{nikolentzos2023weisfeiler} applied the Weisfeiler-Leman algorithm to capture hierarchy in data and built HNNs accordingly. However, none of the above methods have considered comparing distributions before and after embedding into the hyperbolic space.

\subsubsection{GW distance}
GW distance, a variant of optimal transport distance, has been widely utilized to quantify structural differences between different distributions. Memoli provided a detailed introduction and study of GW distance in~\cite{memoli2007use, memoli2011gromov}. Subsequently, Sturm~\cite{sturm2023space} presented a more in-depth mathematical exposition of GW distance, focusing on its geodesic structure and gradient flow. Following these foundational works, GW distance has found applications in various fields, including computer vision~\cite{schmitzer2013modelling, peyre2016gromov}, recommendation systems~\cite{li2022gromov}, natural language processing~\cite{alvarez2018gromov}, generative models~\cite{bunne2019learning, titouan2019sliced, nakagawa2023gromovwasserstein, lee2023monotone}. In particular, Lee \etal~\cite{lee2023monotone} employed the GM formulation to achieve geometry-preserving dimension reduction within a generative modeling framework. The discussion on the GM formulation can also be traced to~\cite{memoli2022distance, dumont2022existence}. Moreover, GW distance has been applied to specific data types such as graphs~\cite{xu2019scalable, xu2019gromov} and specific neural network architectures such as transformers~\cite{huang2022improving}. To the best of our knowledge, GW distance has not yet been explored in the context of hyperbolic neural networks.

\section{Methodology}\label{sec:methodology}
\subsection{Preliminaries of Hyperbolic Geometry}\label{subsec:hyperbolic geometry}
Hyperbolic geometry, characterized by its constant negative curvature, represents a non-Euclidean geometry essential for capturing complex hierarchical representations. We review operations that are defined in two different isometric coordinate systems, namely the Poincar\'{e} ball (Poincar\'{e} disk) model and the Lorentz (hyperboloid) model. 

\subsubsection{Poincar\'{e} ball model}~\cite{ganea2018hyperbolic, shimizu2021hyperbolic}
The Poincar\'{e} ball model for an $n$-dimensional hyperbolic space with curvature $-c$ is defined by
\begin{equation}\label{eq:p_maniflod}
\mathbb{B}_{c}^{n} = \left \{ \rvx\in \mathbb{R}^{n} : c\left \| \rvx \right \| < 1 , c > 0 \right \}, 
\end{equation}
with the corresponding Riemannian metric given by $\mathfrak{g} ^{\mathbb{B}} = (\lambda_{\rvx}^{c})^2 \mI$, where $\lambda _{\rvx}^{c} = {2}({1- c\left \| \rvx \right \|^{2}})^{-1}$ is the conformal factor. 

The analogy of a linear layer in HNN depends on two important operations, namely M\"{o}bius addition and multiplication. The Möbius addition $\oplus $ is defined as:
\begin{equation}\label{p_oplus}
\rvx \oplus_{c} \rvy = \frac{\left (1+2c \left \langle \rvx,\rvy \right \rangle + c \left \| \rvy \right \| ^{2}  \right )\rvx+\left ( 1 - c \left \| \rvx \right \|^{2}   \right )  \rvy  }{1+2c \left \langle \rvx,\rvy \right \rangle + c \left \| \rvx \right \| ^{2}\left \| \rvy \right \| ^{2} }.
\end{equation}
The Möbius multiplication $\rmM \otimes_c \rvx$ is defined as:
\begin{equation}\label{p_otimes_m}
\rmM \otimes_{c} \rvx = \frac{1}{\sqrt{c} } \tanh \left ( \frac{\left \| \rmM\rvx \right \|}{ \rmM\rvx} \tanh^{-1}(\sqrt{c}\left \| \rvx \right \|) \right ) \frac{\rmM\rvx}{\left \| \rmM\rvx \right \| }.
\end{equation}
Finally, the geodesic distance between two points $\rvx$ and $\rvy$ in the Poincar\'{e} ball can be expressed as:
\begin{equation}\label{eq:p_distance2}
d^{c}_\sB \left ( \rvx,\rvy \right ) = \frac{2}{\sqrt{c}}  \tanh^{-1} \left (\sqrt{c} \left \| -\rvx \oplus_{c} \rvy \right \|  \right ).
\end{equation}
Unless otherwise specified, we follow the common choice to take $c=1$.

\subsubsection{Lorentz model}~\cite{chami2019hyperbolic, chen2021fully} The Lorentz model of an $n$-dimensional hyperbolic space $\mathbb{L} ^{n}$ is a manifold embedded in a $(n + 1)$-dimensional Minkowski space. More specifically, this Minkowski space contains the same points as $\R^{n+1}$, but with an inner product $\left \langle \cdot , \cdot \right \rangle _{\mathbb{L}}$ defined by
\begin{equation}\label{eq:L_inner}
\left \langle \rvx,\rvy \right \rangle _{\mathbb{L}} = -x_0 y_0 + \sum_{i=1}^{n} x_i y_i, \rvx=(x_0,\cdots,x_n), \rvy=(y_0,\cdots,y_n)\in \mathbb{R}^{n+1}.
\end{equation}
The Lorentz model $\mathbb{L} ^{n}$ contains points with $\ip{\rvx}{\rvx}_\sL = -c$, where $-c$ is the curvature. That is,
\begin{equation}\label{eq:L_maniflod}
\mathbb{L}^{n} = \left \{ \rvx = (x_0,\cdots , x_n)\in \mathbb{R}^{n+1} : \left \langle \rvx, \rvx \right \rangle _{\mathbb{L}} = -c, x_0 > 0 \right \}, 
\end{equation}
The geodesic distance between two points $\rvx$ and $\rvy$ in the Lorentz model $\mathbb{L}^{n}$ is expressed as
\begin{equation}\label{eq:l_distance}
d^c_\sL \left ( \rvx,\rvy \right ) = \frac{1}{\sqrt{c}} \cosh^{-1}\left (- \left \langle \rvx,\rvy \right \rangle _{\mathbb{L}} \right ). 
\end{equation}

\begin{figure}[!t]
  \centering
  \includegraphics[width=3.75in]{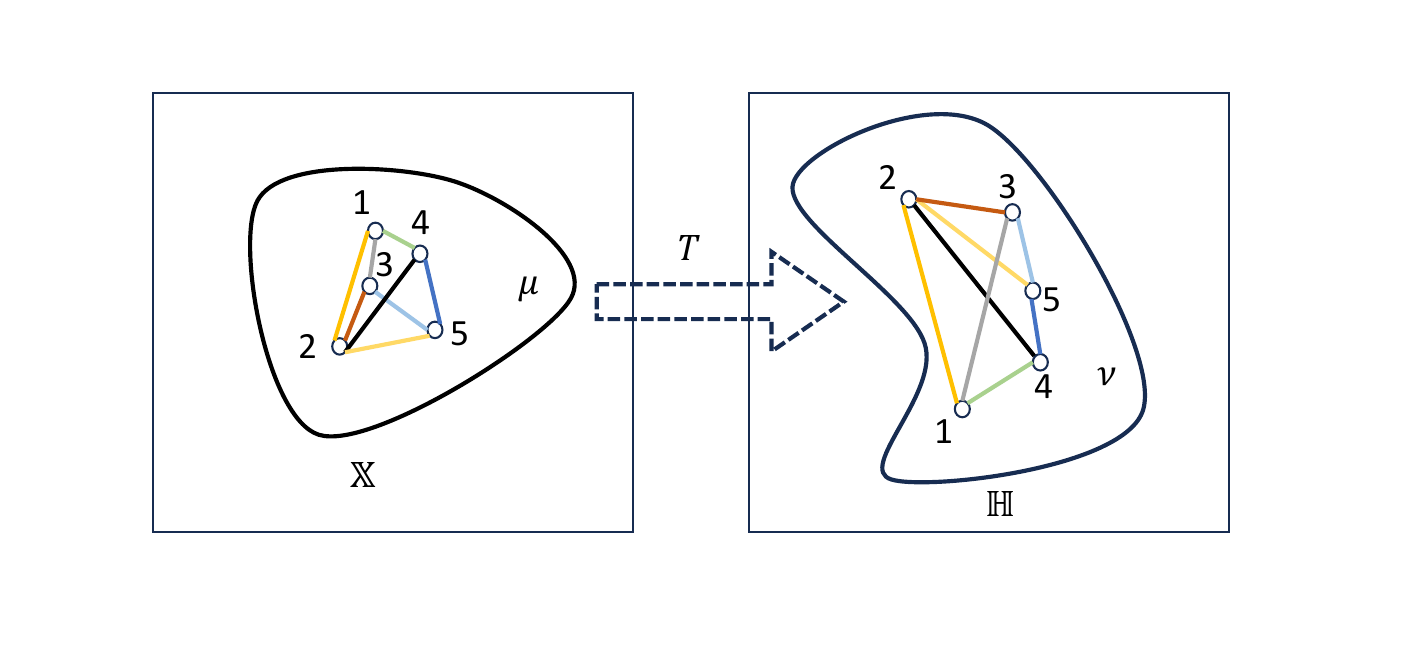}
  \caption{Illustration of the transport map $T$. The feature distribution $\mu$ defined on $\sX$ is pushed forward to $\nu = T _\# \mu$ on $\sH$.}
  \label{GW_fig}
\end{figure}

\subsection{GW Regularization}\label{subsec:GW_distance}
To effectively utilize HNNs for learning from data, extending the geometry-preserving embedding technique introduced in~\cite{lee2023monotone}, we employ a regularized learning approach focused on preserving the intrinsic geometric structures of the input features onto hyperbolic spaces. This involves employing a cost function capable of comparing two measures across distinct geometric spaces.

To bridge this gap, we consider the Gromov-Wasserstein (GW) distance~\cite{memoli2007use} as it is adept at quantifying the similarity between probability distributions defined on distinct metric spaces. For two probability distributions $\mu$ and $\nu$ in a Euclidean feature space $\sX$ and a hyperbolic space $\sH$, respectively, the GW distance is defined as:
\begin{align}
\operatorname {GW} \left ( \mu ,\nu  \right ) : 
&:= \min_{\pi\in \Pi \left(  \mu ,\nu \right ) } \underset{\left ( (\rvx,\rvy ),({\rvx}',{\rvy}' ) \right )\sim \pi^{2}  }{\mathbb{E}}  \left [ \left | c_\sX(\rvx,{\rvx}' )- c_\sH(\rvy,{\rvy}' ) \right |^{2}  \right ] \label{eq:gw_distance2}
\end{align}
where $c_\sX$ and $c_\sH$ represent the cost functions in the two spaces, and $\Pi (\mu, \nu)$ denotes the set of transport plans between $\mu$ and $\nu$, which contains all joint distributions whose first and second marginals are given by $\mu$ and $\nu$, respectively. By comparing the pairwise distances between two probability distributions to evaluate the similarity of geometric structures across different metric spaces.

When working with HNNs, an explicit map $T$ from $\sX$ to $\sH$ is induced by the HNN layers, as illustrated in Fig.~\ref{GW_fig}. The minimization problem presented in~\eqref{eq:gw_distance2} can be reformulated as a Gromov-Monge (GM) distance minimization problem concerning a map $T$, defined as
\begin{equation}\label{eq:gm_distance2}
\operatorname {GM} \left ( \mu ,\nu  \right ) : = \min_{T_{\#}\mu = \nu} \underset{(\rvx,{\rvx}')\sim \mu^{2}  }{\mathbb{E}}  \left [ \left | c_\sX(\rvx,{\rvx}' )- c_\sH(T(\rvx),T({\rvx}') ) \right |^{2}  \right ]  .
\end{equation}

In an HNN, when embedding the probability distribution $\mu$ from a Euclidean space to a hyperbolic space, we can consider minimizing the following cost over ${T \in \gH}$, where $\gH$ is a hypothesis class determined by the HNN:
\begin{equation}\label{eq:GM_distr}
    \operatorname{GM}(T; \mu):= \underset{(\rvx,{\rvx}')\sim \mu^{2}  }{\mathbb{E}}  \left [ \left | c_\sX(\rvx, \rvx')- c_\sH(T(\rvx),T({\rvx}') ) \right |^{2}  \right ].
\end{equation}
Thus, we have $\min_{T \in \gH} \operatorname{GM}(\mu, T_\#\mu) = \min_{T \in \gH} \operatorname{GM}(T; \mu)$. While the GM problem described in~\eqref{eq:gm_distance2} is highly challenging due to its pushforward constraint, minimizing the cost function in~\eqref{eq:GM_distr} is comparatively easier, as it does not require the pushforward constraint.

In practical implementation, we can employ the simple cost function $c_\sX(\rvx,{\rvx}') = \left \| \rvx- {\rvx}' \right \|^{2}$ or other cost functions such as $c_\sX(\rvx,{\rvx}') = \log \left( 1 + \left \| \rvx- {\rvx}' \right \|^{2} \right)$. Additionally, the explicit form of $c_\sH$ can be derived from equations \eqref{eq:p_distance2} and \eqref{eq:l_distance}: we can use $c_\sH = d^{c}_\sH$ or $c_\sH = \log \left( 1 + d^{c}_\sH \right)$, where $\sH = \sB$ or $\sH = \sL$. 

When we have data points $\{\rvx_i\}_{i=1}^m$ sampled independently and identically distributed (i.i.d.) from $\mu$, we consider an empirical version of the GM distance, namely,
\begin{equation}\label{eq:GM_points}
    {\operatorname{GM}}(T; \{\rvx_i\}_{i=1}^m) = \frac{1}{m(m-1)} \sum_{i,j=1}^m \abs{ c_\sX(\rvx_i , \rvx_j) - c_\sH(T(\rvx_i), T(\rvx_j)) }^2.
\end{equation}

\begin{figure}[!t]
  \centering
  \includegraphics[width=\textwidth]{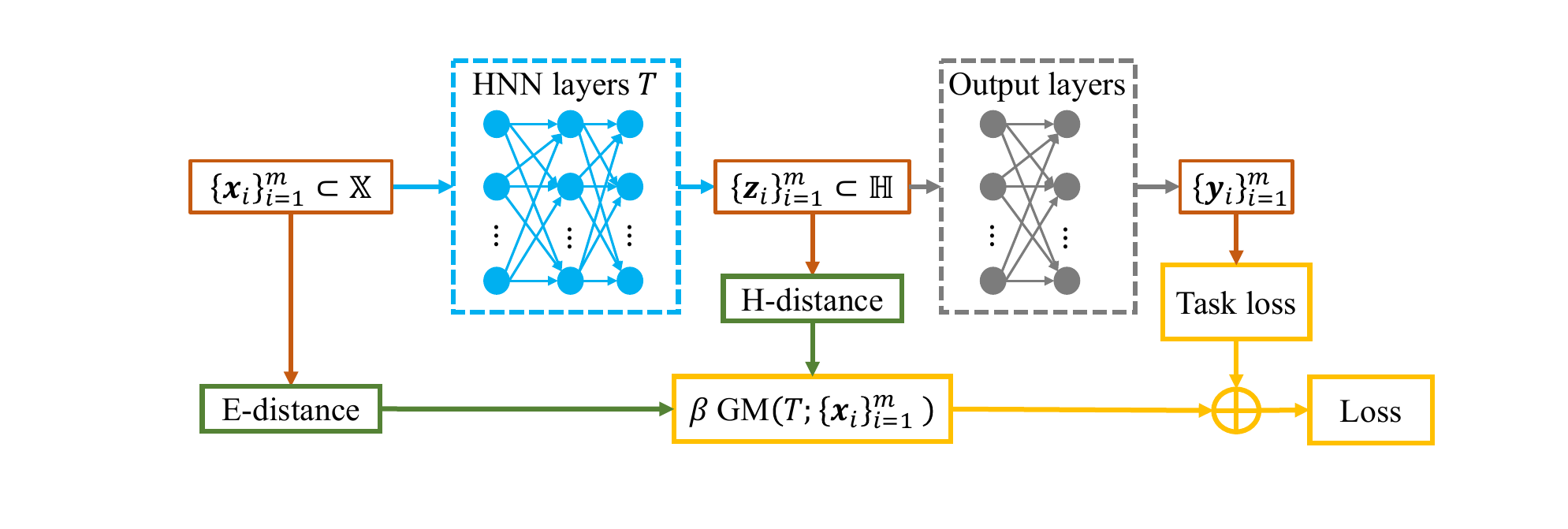}
  \caption{The framework of our GW regularization.}
  \label{regularization_gw}
\end{figure}

The overall framework of our GW regularization is shown in Fig.~\ref{regularization_gw}. Here, $\{\rvx_i\}_{i=1}^m$ are the Euclidean features that are fed into the HNN layers denoted by $T$, $\{\rvz_i\}_{i=1}^m$ are the hyperbolic representations so that $\rvz_i = T(\rvx_i)$ for each $i$. The GM distance, calculated according to \eqref{eq:GM_points}, is used as a penalty term, multiplied with a hyperparameter $\beta$ and then added to the loss function associated with the specific task. 

\subsubsection{Generalization analysis}
In the tasks we consider, the ability to generalize beyond the given data points is essential. Although \eqref{eq:GM_points} only provides a formulation based on the training data, we anticipate that it will serve as an effective approximation of \eqref{eq:GM_distr}, provided that the training set $\{\rvx_i\}_{i=1}^m$ adequately captures the characteristics of the underlying distribution $\mu$. This expectation is encapsulated in the theorem we present below, which formalizes the relationship between the pointwise formulation and its distributional approximation under the condition of a representative sample.

\begin{theorem}\label{thm}
Given cost functions $c_\sX$ and $c_\sH$, suppose that there exists a constant $0<\alpha\leq1$ for which $T$ satisfies the following bi-Lipschitz condition:
\begin{equation}\label{eq:bilip}
    \alpha c_\sX({\rvx,\rvx'}) \leq c_\sH(T(\rvx), T(\rvx')) \leq \frac{1}{\alpha} c_\sX({\rvx,\rvx'}),
\end{equation}
for any $\rvx, \rvx' \in \sX$. Let $\mu$ be a distribution defined on $\sX$ and $\{\rvx_i\}_{i=1}^m$ be i.i.d. sampled from $\mu$. Then,
\begin{align}\label{eq:bilip_res}
    \left| \operatorname{GM}(T; \{\rvx_i\}_{i=1}^m) - \operatorname{GM}(T;\mu) \right| \leq C \frac{(1/\alpha-1)^2}{R^2}
\end{align}
holds with probability at least $\displaystyle 1 - 2\exp\left(- \frac{m C}{8R^4}\right)$
where $C$ is a constant depending on $\mu$ and $R:=\max_{\rvx,\rvx'\in  \mathrm{supp}(\mu)} c_\sX(\rvx,\rvx')$.
\end{theorem}

The proof of Theorem~\ref{thm} is presented in the appendix.

\subsubsection{Complexity analysis}
Since the time complexity for calculating each Euclidean distance is $O(n)$ and the time complexity for calculating each hyperbolic distance (in either the Poincar\'{e} ball model or the Lorentz model) is $O(n)$, the time complexity for calculating the GW regularization term is $O(nm^2)$. 

Note that, if we only consider linear layers in an $L$-layer Vanilla HNN, the time complexity is $O(mn^2L)$. In the few-shot learning task and the semi-supervised graph node-level tasks that we consider in this paper, the number of data points $m$ is usually smaller than or comparable with $n$. Moreover, in tasks such as link prediction, one also needs to compare pairs of data points, leading to a time complexity of $O(mn^2L)$ in computing the task loss. Therefore, adding the regularization term will not lead to change of the order of time complexity.

\section{Experiments}
\label{sec:Experiment}
To validate the effectiveness of the GW regularization method, we conduct experiments on both image datasets and graph datasets. We present results on few-shot image classification in Section~\ref{subsec:Experiment_image} and results on graph node classification and link prediction in Section~\ref{subsec:Experiment_graph}. All the presented experiments are implemented on a server with RTX 4090 (24GB) GPUs, where each implementation runs on a single GPU.
The implementation of our proposed methods can be accessed at \url{https://github.com/yyf1217/GW-Regularization}. 

\subsection{Few-Shot Image Classification}\label{subsec:Experiment_image}
We consider the important task of few-shot image classification where the goal is to recognize new categories with very few labeled examples per class. Unlike supervised image classification, this task aims to generalize from a small number of examples, typically one to five images per class. Learning structures of images is crucial since there is very limited labeled data.

\subsubsection{Datasets}
We consider two widely used image datasets, namely \emph{MiniImageNet} \cite{vinyals2016matching} and \emph{Caltech-UCSD Birds-200-2011 (CUB)} \cite{wah2011caltech}. These datasets have been used in \cite{khrulkov2020hyperbolic}, where the hyperbolic ProtoNet model excels Euclidean baselines. The {MiniImageNet} dataset is a subset of the ImageNet dataset \cite{deng2009imagenet}. It comprises images from 100 classes, with 600 images per class. The classes have a 64/16/20 split for training/validation/test. The {CUB} dataset consists of 200 categories of bird images, totaling 11,788 pictures. In this dataset, the training/validation/test split is 100/50/50.

\subsubsection{Setting}
We apply our GW regularization on the hyperbolic ProtoNet model \cite{khrulkov2020hyperbolic}, which primarily consists of convolutional neural network (CNN) layers used for feature extraction from Euclidean image data and subsequent HNN layers for classification. In \cite{khrulkov2020hyperbolic}, the HNN layers are taken to be a simple exponential map from the Euclidean space to the Poincar\'{e} ball. In our implementation, we utilize the publicly available code for the hyperbolic ProtoNet model from \url{https://github.com/leymir/hyperbolic-image-embeddings}.

\begin{figure}[!t]
  \centering
  \includegraphics[width=\textwidth]{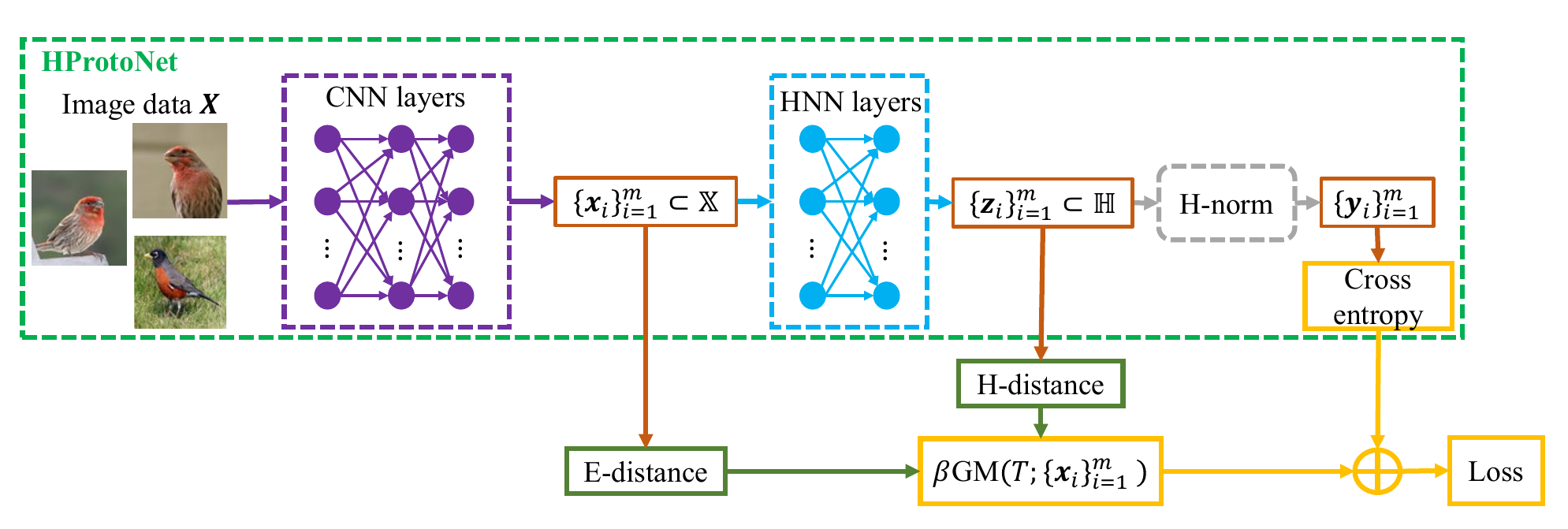}
  \caption{The framework of the GW regularized hyperbolic ProtoNet model.}
  \label{HProtoNet_gw}
\end{figure}

For clarity, we illustrate the framework for this task in Fig.~\ref{HProtoNet_gw}, which is based on Fig.~\ref{regularization_gw}. We take the features $\{\rvx_i\}_{i=1}^m$ from the output of the CNN layers, and then the HNN layers process them into $\{\rvz_i\}_{i=1}^m$. Following this, the hyperbolic distance between each hyperbolic embedding $\rvz_i$ and the origin of the Poincar\'{e} ball is computed according to~\eqref{eq:p_distance2}, which is then used to calculate the maximum class probability. Finally, the cross-entropy loss is computed using the maximum class probability and the labels. We add $\beta \operatorname{GM} (T; \{\rvx_i\}_{i=1}^m)$ to the cross-entropy loss to formulate our regularized loss function.

We consider various CNN architectures including Conv4, ResNet10, ResNet12, and ResNet18. The HNN layers within the model incorporate linear layers from \cite{ganea2018hyperbolic}. Additionally, ReLU is used as the activation function across the model. For Conv4, the embedding dimension is set to 1,600, while for ResNet10, ResNet12, and ResNet18, the embedding dimension is set to 512. The initial learning rate is 0.001. 
We consider both the 1-Shot 5-Way and 5-Shot 5-Way settings, which means there are five classes involved for the task and only one/five examples per class for training. In all experiments for the 1-Shot 5-Way task, a consistent curvature parameter of $-0.08$ is employed. Similarly, for the 5-Shot 5-Way task, we uniformly set the curvature to $-0.01$ throughout all experiments. Both curvatures agree with the recommened settings in the original hyperbolic ProtoNet. The hyperparameter $\beta$ is selected according to the validation set. We report the initial learning rate as well as the best $\beta$ in Table~\ref{tab:HPr_hyperparameter}.

\begin{table}[t]
\scriptsize
\caption{Hyperparameters used in few-shot image classification. ``ResNet'' abbreviated as ``Res'' due to space constraints.}
\label{tab:HPr_hyperparameter}
\begin{center}
\renewcommand\arraystretch{1.25}
\begin{tabular}{lccccccccc}
    \toprule
   & & \multicolumn{4}{c}{Initial learning rate (lr)} & \multicolumn{4}{c}{$\beta$}       \\
   & & Conv4  & Res10  & Res12 & Res18  & Conv4 & Res10  & Res12 & Res18\\
    \midrule
\multirow{2}{*}{MiniImageNet}   & $1$-Shot $5$-Way &  $0.0005$ & $0.001$ & $0.001$ & $0.001$ & $0.6$ & $0.5$ & $0.1$ & $0.7$ \\
  & $5$-Shot $5$-Way   & $0.01$ & $0.001$ & $0.001$ & $0.001$ & $14$ & $0.1$ & $0.5$ & $0.5$ \\
\midrule
\multirow{2}{*}{CUB}   & $1$-Shot $5$-Way    &  $0.001$ & $0.001$ & $0.001$ & $0.001$ & $0.02$ & $0.015$ & $0.15$ & $0.1$ \\
  & $5$-Shot $5$-Way     &  $0.01$ & $0.001$ & $0.001$ & $0.001$  & $0.07$ & $0.01$ & $0.25$ & $0.1$ \\
\bottomrule
\end{tabular}
\end{center}
\end{table}

\subsubsection{Results}
We present our numerical results for both the 1-Shot 5-Way and 5-Shot 5-Way settings in Table~\ref{tab:gw_image_sult}. We randomly sample 10,000 data points from the test set to evaluate the performance of the model and report 95\% confidence intervals for all results. Each entry reports the average classification accuracy, including means and standard deviations.

\begin{table}[htbp]
\caption{Accuracy (\%) results for few-shot image classification.}
\label{tab:gw_image_sult}
\begin{center}
\begin{tabular}{lcccc}
    \toprule
    & \multicolumn{2}{c}{MiniImageNet} & \multicolumn{2}{c}{CUB}     \\
    & $1$-Shot $5$-Way & $5$-Shot $5$-Way  & $1$-Shot $5$-Way & $5$-Shot $5$-Way  \\
    \midrule
Conv4  & $53.77{\pm 0.20}$   & $71.33{\pm 0.16}$     & $64.66{\pm 0.23}$   & $80.29{\pm 0.16}$   \\
Conv4+GW     & $55.57{\pm 0.22}$   & $73.04{\pm 0.16}$  & $68.67{\pm 0.23}$& $80.54{\pm 0.16}$   \\
\hline
ResNet10  & $56.45{\pm 0.21}$   & $65.39{\pm 0.17}$     & $72.01{\pm 0.22}$   & $80.75{\pm 0.15}$   \\
ResNet10+GW     & $57.50{\pm 0.22}$   & $67.62{\pm 0.17}$  & $72.67{\pm 0.22}$& $83.73{\pm 0.14}$   \\
\hline
ResNet12  & $55.96{\pm 0.22}$   & $70.32{\pm 0.17}$     & $75.77{\pm 0.22}$   & $82.90{\pm 0.15}$   \\
ResNet12+GW     & $56.95{\pm 0.22}$   & $73.04{\pm 0.16}$  & $77.24{\pm 0.21}$& $85.66{\pm 0.14}$   \\
\hline
ResNet18  & $57.04{\pm 0.22}$   & $68.43{\pm 0.16}$     & $71.86{\pm 0.22}$   & $85.31{\pm 0.13}$   \\
ResNet18+GW     & $58.51{\pm 0.22}$   & $71.64{\pm 0.16}$  & $72.64{\pm 0.22}$& $85.83{\pm 0.13}$   \\
  \bottomrule
\end{tabular}
\end{center}
\end{table}

From the results, we observe that our GW regularization consistently boosts the accuracy of the model. This improvement is evident in both datasets and both scenarios, regardless of the underlying CNN architecture. For most CNN architectures, the improvement is more evident in the 5-Shot 5-Way scenario than in the 1-Shot 5-Way scenario. We believe that this can be attributed to the increased number of training instances available in the 5-Shot scenario, which provides a richer structure that GW regularization can leverage.

\subsubsection{Runtime}
We report the average run time for training one epoch in Table~\ref{tab:gw_image_time}. As anticipated, incorporating GW regularization results in extended epoch training times. However, the increase in runtime is not significant, aligning with the time complexity analysis in Section~\ref{subsec:GW_distance}. This observation underscores the efficiency of our GW regularization approach, ensuring that the added computational demand does not impose a significant burden on the overall training process.

\begin{table}[htbp]
\caption{Average runtime (in seconds) for training one epoch in few-shot image classification.}
\label{tab:gw_image_time}
\begin{center}
\begin{tabular}{lcccc}
    \toprule
    & \multicolumn{2}{c}{MiniImageNet} & \multicolumn{2}{c}{CUB}     \\
    & $1$-Shot $5$-Way & $5$-Shot $5$-Way  & $1$-Shot $5$-Way & $5$-Shot $5$-Way  \\
    \midrule
Conv4  & $31.2$   & $25.2$     & $32.4$   & $32.9$   \\
Conv4+GW     & $33.6$   & $26.4$  & $35.1$& $32.9$   \\
\hline
ResNet10  & $37.2$   & $28.8$     & $38.4$   & $34.9$   \\
ResNet10+GW     & $40.8$   & $34.8$  & $43.2$& $38.4$   \\
\hline
ResNet12  & $42.0$   & $36.0$     & $46.8$   & $42.0$   \\
ResNet12+GW     & $45.6$   & $40.8$  & $50.4$& $45.6$   \\
\hline
ResNet18  & $55.2$   & $43.2$     & $60.0$   & $52.8$   \\
ResNet18+GW     & $57.6$   & $49.2$  & $62.4$& $57.6$   \\
  \bottomrule
\end{tabular}
\end{center}
\end{table}

\subsection{Semi-Supervised Link Prediction and Node Classification}\label{subsec:Experiment_graph}
We consider graph link prediction and node classification, which are widely benchmarked semi-supervised tasks in graph deep learning literature.

\subsubsection{Datasets}
We consider nine datasets for node classification, namely \emph{Cora, Disease, Airport, Cornell, Texas, Wisconsin, Chameleon, Squirrel, and Actor}. For the first three datasets, namely Disease, Airport, and Cora, we also perform link prediction. We briefly introduce the datasets in the appendix and summarize the statistics of the graphs in Table~\ref{tab:gw_dataset}.

\begin{table}[h!]
\scriptsize
  \caption{Statistics of graph datasets.}
  \label{tab:gw_dataset}
  \centering
  \begin{tabular}{@{}lcccccccccc@{}}
    \toprule
    & Disease-LP & Disease-NC & Airport & Cora & Cornell    & Texas      & Wisconsin  & Chameleon  & Squirrel   & Actor       \\
    \midrule
    \# nodes & 2,665 & 1,044      & 3,188 & 2,708 & 183 & 183      & 251 & 2,277 & 5,201 & 7,600 \\
    \# edges & 2,265        & 2,265             & 15,837        & 4,488 & 280 & 295      & 466 & 31,421 & 198,493 & 26,752        \\
    \# features     & 11        & 1,000            & 11        & 1,433     & 1,703 & 1,703      & 1,703 & 2,325 & 2,089 & 931   \\
    \# classes     & N/A         & 2            & 4        & 7    & 5 & 5      & 5 & 5 & 5 & 5   \\
  \bottomrule
  \end{tabular}
\end{table}

In link prediction tasks, we randomly split edges into $85\%/5\%/10\%$ for training, validation, and test sets. For node classification tasks, we use a $70\%/15\%/15\%$ splits for Airport, Cornell, Texas, Wisconsin, Chameleon, Squirrel, and Actor, we use $30\%/10\%/60\%$ splits for Disease, and we use standard splits with $20$ train examples per class for Cora. The above split settings agree with standard settings in the baseline papers.

\subsubsection{Setting}
For the graph datasets, we consider the following three HNNs: the vanilla HNN~\cite{ganea2018hyperbolic}, HGCN~\cite{chami2019hyperbolic}, and HyboNet~\cite{chen2021fully}. We utilize the publicly available code for HNN and HGCN from \url{https://github.com/HazyResearch/hgcn} and HyboNet from \url{https://github.com/chenweize1998/fully-hyperbolic-nn}. 

For each model, we apply GW distance to the input node features and their hyperbolic representations after applying the Euclidean-to-hyperbolic operation. The regularization framework is the same as Fig.~\ref{regularization_gw}.

In these experiments, the HNN models share a common architecture consisting of HNN layers and output layers. These models employ different HNN layers to extract features from graph data and generate hyperbolic embeddings $\{\rvz_i\}_{i=1}^m$. Specifically, the Vanilla HNN and HGCN employ the Poincar\'{e} model to build linear layers and activation. HGCN also has aggregation operation, facilitating message passing. Differently, HyboNet employs the linear layers and aggregation in the Lorentz model. The output layers are used to accomplish various tasks. For node classification, the obtained hyperbolic embeddings are directly projected to the Euclidean space. A linear layer is then used to predict the class probabilities, with which the negative log-likelihood loss is computed. For link prediction, the hyperbolic distances between the hyperbolic embeddings are computed. Subsequently, the Fermi-Dirac algorithm \cite{krioukov2010hyperbolic} is employed to transform the hyperbolic distances into the edge probabilities. Finally, cross-entropy losses are computed using these edge probabilities and the true edge labels.

For simplicity, the curvature of the hyperbolic space is set as $-1$ for all experiments, which is a common setting for graph datasets. Tables~\ref{tab:hnn_hgcn_hyp} and \ref{tab:hybonet_hyp} list hyperparameters such as initial learning rate, the number of the HNN layers, weight decay, dropout, and embedding dimensions. The hyperparameter $\beta$ from validation is also listed for different datasets.

\begin{table}[h!]
\scriptsize
  \caption{Hyperparameters used in the Vanilla HNN and HGCN.}
  \label{tab:hnn_hgcn_hyp}
  \centering
  \begin{tabular}{@{}lcccccccccccc@{}}
    \toprule
     & \multicolumn{2}{c}{Disease} & \multicolumn{2}{c}{Airport} & \multicolumn{2}{c}{Cora} & Cornell    & Texas      & Wisconsin  & Chameleon  & Squirrel   & Actor       \\
           & LP         & NC             & LP         & NC             & LP         & NC          & NC         & NC         & NC         & NC         & NC         & NC          \\
    \midrule
Initial lr       & $0.01$ & $0.01$  & $0.01$     & $0.01$     & $0.01$     & $0.01$     & $0.01$    & $0.01$     & $0.01$     & $0.01$     & $0.01$     & $0.01$  \\
\# layers      & $2$        & $2$     & $2$     & $2$     & $ 2$     & $ 2$     & $2$     & $2$     & $ 2$     & $2$     & $2$     & $2$  \\
Weight decay     & $0.0$        & $0.001$     & $0.0005$     & $0.0$     & $0.005$     & $0.001$     & $0.001$     & $0.001$     & $0.001$     & $0.001$     & $0.001$     & $0.001$  \\
Dropout       & $0.0$        & $0.1$   & $0.0$     & $0.0$     & $0.5$     & $0.5$     & $0.5$     & $0.5$     & $0.5$     & $0.5$   & $0.5$     & $0.5$  \\
\# embeddings    & $3$        & $16$     & $16$     & $16$     & $ 16$     & $ 16$     & $16$     & $16$     & $ 16$     & $ 16$     & $ 5$     & $ 5$  \\
HNN $\beta$      & $0.9$       & $0.08$  & $ 2$     & $1.25$     & $ 0.5$     & $ 0.35$     & $0.25$     & $0.1$     & $ 0.15$     & $ 0.02$     & $0.001 $     & $ 0.03$  \\
HGCN $\beta$      & $0.3$       & $0.09$  & $ 2$     & $1.5$     & $ 0.1$     & $ 0.6$     & $0.25$     & $0.15$     & $ 0.2$     & $ 0.08$     & $0.01 $     & $ 0.03$  \\
  \bottomrule
  \end{tabular}
\end{table}

\begin{table}[h!]
\scriptsize
  \caption{Hyperparameters used in HyboNet.}
  \label{tab:hybonet_hyp}
  \centering
  \begin{tabular}{@{}lcccccccccccc@{}}
    \toprule
     & \multicolumn{2}{c}{Disease} & \multicolumn{2}{c}{Airport} & \multicolumn{2}{c}{Cora} & Cornell    & Texas      & Wisconsin  & Chameleon  & Squirrel   & Actor       \\
           & LP         & NC             & LP         & NC             & LP         & NC          & NC         & NC         & NC         & NC         & NC         & NC          \\
    \midrule
Initial lr       & $0.005$ & $0.005$  & $0.01$     & $0.02$     & $0.02$     & $0.02$     & $0.005$    & $0.005$     & $0.005$     & $0.005$     & $0.01$     & $0.01$  \\
\# layers      & $2$        & $4$     & $2$     & $6$     & $ 2$     & $ 3$     & $2$     & $2$     & $ 2$     & $2$     & $2$     & $2$  \\
Weight decay     & $0.0$        & $0.0$     & $0.0$     & $0.0001$     & $0.001$     & $0.01$     & $0.0$     & $0.0$     & $0.0$     & $0.0$     & $0.001$     & $0.001$  \\
Dropout       & $0.0$        & $0.1$   & $0.0$     & $0.0$     & $0.7$     & $0.9$     & $0.2$     & $0.2$     & $0.2$     & $0.2$   & $0.2$     & $0.2$  \\
\# embeddings      & $16$        & $16$     & $16$     & $16$     & $ 16$     & $ 16$     & $16$     & $16$     & $ 16$     & $ 16$     & $ 16$     & $ 16$  \\
$\beta$      & $0.3$       & $0.07$  & $ 1.5$     & $1.1$     & $ 0.25$     & $ 0.25$     & $0.13$     & $0.1$     & $ 0.1$     & $ 1.3$     & $2.5 $     & $ 0.01$  \\
  \bottomrule
  \end{tabular}
\end{table}

\subsubsection{Results}
For the link prediction task, we report the AUC scores in Table~\ref{tab:gw_lp_result}. For the node classification task, we report the F1 scores, in Table~\ref{tab:gw_nc_result1}. Each score reports the average from three random runs as well as the standard deviation. For clarity, we just use ``HNN'' in place of the Vanilla HNN in the tables.

\begin{table}[h!]
  \caption{AUC (\%) results of the link prediction task.}
  \label{tab:gw_lp_result}
  \centering
  \begin{tabular}{@{}lccc@{}}
    \toprule
     & Disease & Airport & Cora        \\
    \midrule
HNN        & $92.83{\pm1.83}$ & $93.56{\pm0.30}$  & $88.80{\pm1.29}$  \\
HNN+GW     & $94.45{\pm0.58}$ & $94.23{\pm0.33}$  & $89.76{\pm2.07}$ \\
\hline
HGCN       & $92.41{\pm1.78}$ & $93.42{\pm0.15}$  & $93.37{\pm0.15}$ \\
HGCN+GW    & $ 93.79{\pm1.15 }$ & $ 95.81{\pm0.02 }$ & $ 93.48{\pm0.22 }$  \\
\hline
HYBONET    & $ 95.65{\pm0.57 }$ & $ 96.18{\pm0.05 }$ & $ 92.04{\pm0.37 }$ \\
HYBONET+GW & $ 96.58{\pm0.57 }$ & $ 96.44{\pm0.02 }$  & $ 93.33{\pm0.22 }$ \\
  \bottomrule
  \end{tabular}
\end{table}

\begin{table}[h!]
  \caption{F1 (\%) result of the node classification task.}
  \label{tab:gw_nc_result1}
  \centering
  \begin{tabular}{@{}lccccc@{}}
    \toprule
    & Disease & Airport & Cora  & Chameleon  & Squirrel    \\
    \midrule
HNN      & $58.40{\pm2.53 }$     & $87.40{\pm1.34 }$   & $53.73{\pm0.91}$ & $71.43{\pm1.63}$     & $47.89{\pm2.21}$   \\
HNN+GW   & $65.62{\pm1.98}$     & $89.38{\pm0.67}$     & $55.33{\pm0.35}$ & $72.53{\pm0.19}$     & $49.44{\pm1.05}$   \\
\hline
HGCN     & $93.04{\pm1.64}$     & $86.96{\pm0.22}$     & $79.47{\pm1.01}$  & $77.72{\pm0.90}$   & $51.79{\pm0.58}$   \\
HGCN+GW   & $ 95.28{\pm0.40}$     & $ 87.91{\pm0.72}$     & $ 80.00{\pm0.62  }$  & $ 79.18{\pm1.64 }$  & $ 52.70{\pm1.24}$  \\
\hline
HYBONET   & $ 85.70{\pm0.60}$     & $ 93.13{\pm0.19}$     & $ 73.53{\pm0.75}$   & $ 82.05{\pm0.55}$  & $ 74.04{\pm2.42}$  \\
HYBONET+GW  & $ 87.66{\pm1.21 }$     & $ 94.46{\pm0.20}$     & $ 79.40{\pm0.95 }$ & $ 83.58{\pm0.28}$  & $ 77.27{\pm0.78}$ \\
  \bottomrule
  \end{tabular}

  \centering
  \begin{tabular}{@{}l@{\hspace{1.75cm}}cccc@{}}
    \toprule
& Cornell    & Texas  & Wisconsin  & Actor       \\
    \midrule
HNN        & $93.94{\pm1.31}$     & $94.69{\pm1.31}$ & $93.21{\pm3.86}$    & $42.63{\pm1.73}$  \\
HNN+GW   & $95.45{\pm2.28}$     & $96.97{\pm1.32}$  & $98.15{\pm1.85}$      & $45.03{\pm1.73}$  \\
\hline
HGCN  & $83.33{\pm3.47}$     & $78.79{\pm4.73}$    & $75.31{\pm2.83}$   & $36.39{\pm1.00}$  \\
HGCN+GW      & $ 88.64{\pm2.28 }$     & $ 81.82{\pm3.94 }$   & $ 83.33{\pm1.86 }$  & $ 39.19{\pm0.67}$  \\
\hline
HYBONET     & $ 78.79{\pm4.73 }$     & $ 67.42{\pm3.47 }$   & $ 74.07{\pm1.86 }$  & $ 45.74{\pm2.13}$  \\
HYBONET+GW  & $ 81.82{\pm3.93 }$     & $ 74.24{\pm1.31 }$    & $ 80.87{\pm3.86 }$ & $ 49.64{\pm0.38 }$ \\
  \bottomrule
  \end{tabular}
\end{table}

From Table~\ref{tab:gw_lp_result}, it is evident that our GW regularization  consistently enhances the AUC scores across different network architectures and across all datasets. Similarly, Table~\ref{tab:gw_nc_result1} showcases that GW regularization again demonstrates a positive impact on model performance across various datasets and architectures. In many scenarios of node classification, GW regularization actually achieves substantial improvement, such as HNN+GW for Disease, HGCN+GW for Cornell and HyboNet+GW for Wisconsin.

\subsubsection{Runtime}
We report the average run time for training one epoch for link prediction in Table~\ref{tab:gw_lp_time}. Similarly to the previous experiment, the runtime for GW regularized training is longer, but not significantly. This again aligns with the complexity analysis in Section~\ref{subsec:GW_distance}.

\begin{table}[h!]
  \caption{Average runtime (in seconds) for training one epoch in link prediction.}
  \label{tab:gw_lp_time}
  \centering
  \begin{tabular}{@{}lccc@{}}
    \toprule
     & Disease & Airport & Cora        \\
    \midrule
HNN        & $ 0.0301$ & $0.0881$& $0.0549$  \\
HNN+GW     & $0.0313$ & $0.0884$  & $0.0563$ \\
\hline
HGCN       & $0.0489$ & $0.1075$  & $0.0743$ \\
HGCN+GW    & $0.0505$ & $ 0.1078$ & $0.0747$  \\
\hline
HYBONET    & $0.0301$ & $ 0.0631$ & $ 0.0374$ \\
HYBONET+GW & $0.0323$ & $ 0.0639$  & $0.0379$ \\
  \bottomrule
  \end{tabular}
\end{table}

\section{Conclusion}

In this paper, we have delved into the integration of the GW distance as a novel regularization term within the realms of hyperbolic neural networks, with a particular emphasis on leveraging the GM formulation. This approach has allowed for a sophisticated comparison of distributions across the Euclidean space and the hyperbolic space, capturing the essence of the underlying structures while not increasing the order of the time complexity. We have demonstrated that the hyperbolic embeddings of the data points resonate closely with distributions where they are sampled from. 

Our regularization method has been rigorously tested and validated across diverse datasets including image data for the few-shot classification task and graph data for the semi-supervised link prediction and node classification tasks, showcasing a uniform elevation in performance metrics.

Our future work involves more efficient methods in large scale settings, where the number of instances is much larger than feature dimensions. Additionally, we are intrigued by the potential of integrating our GW regularization framework with emerging geometric deep learning architectures for complex data structures, where the flexibility with the choice of underlying spaces is crucial.


\section*{Acknowledgement}
YY and DZ acknowledge funding from National Natural Science Foundation of China (NSFC) under award number 12301117, WL acknowledges funding from the National Institute of Standards and Technology (NIST) under award number 70NANB22H021, and GL acknowledges funding from NSF award DMS 2124913.

%
%
\bibliographystyle{splncs04}
\bibliography{egbib}

\newpage
\appendix

\section*{Appendix}
\section{Proof of Theorem~\ref{thm}}
\begin{proof}
    For convenience, we define a binary function $f$ by
    \begin{align*}
        f(\rvx,\rvx') := |c_{\sX}(\rvx,\rvx') - c_{\sH}(T(\rvx),T(\rvx'))|^2.
    \end{align*}
    Then
    \begin{align*}
        \operatorname{GM}(T;\{\rvx_i\}^m_{i=1}) = \frac{1}{m(m-1)}\sum_{i\neq j} f(\rvx_i,\rvx_j).
    \end{align*}
    According to \eqref{eq:bilip}, we upper bound $f$ by
    \begin{align*}
        f(\rvx,\rvx') \leq \left(\frac{1}{\alpha}-1\right)^2 c_{\sX}(\rvx,\rvx')^2 \leq \left(\frac{1}{\alpha}-1\right)^2 R^2 =: b.
    \end{align*}
    Moreover, consider the variance
    \begin{align*}
        \sigma^2 :&= \mathbb{E}\left[\big(f(\rvx_i,\rvx_j) - \mathbb{E}(f(\rvx_i,\rvx_j))\big)^2\right]\\
        &\leq \mathbb{E}[f(\rvx_i,\rvx_j)^2]\\
        &\leq \int |c_{\sX}(\rvx,\rvx') - c_{\sH}(T(\rvx),T(\rvx'))|^4 d\mu d\mu\\
        &\leq \left(\frac{1}{\alpha}-1\right)^4\int c_{\sX}(\rvx,\rvx')^4 d\mu d\mu.
    \end{align*}
    Let $C := \int c_{\sX}(\rvx,\rvx')^4 d\mu d\mu$. By Bernstein's inequality (\eg,~\cite[Theorem 5.15]{calder2020calculus}), 
    \begin{align*}
        \left| \operatorname{GM}(T; \{\rvx_i\}_{i=1}^m) - \operatorname{GM}(T,\mu) \right| \leq t
    \end{align*}
    holds with probability at least $\displaystyle 1-2\exp\left(- \frac{mt^2}{6\left(\sigma^2 + \frac{1}{3}bt\right)}\right)$.
    By letting $t = C (1/\alpha-1)^2/R^2,$
    this reduces to the conclusion of Theorem \ref{thm}, that is,  \eqref{eq:bilip_res} holds with probability at least $1 - 2\exp\left(- \frac{m C}{8R^4}\right)$.
\end{proof}

\section{Introduction of Graph Datasets}

We provide a brief description of the graph datasets used in Section~\ref{subsec:Experiment_graph}.

\paragraph{Disease\cite{chami2019hyperbolic}} This dataset models the SIR epidemiological process, depicting individuals as nodes and their interactions as edges. Nodes carry attributes indicating susceptibility to infection, whereas node classifications reflect the current infection status. Two distinct subsets, Disease-LP and Disease-NC, facilitate link prediction and node classification studies, respectively.
\paragraph{Airport\cite{chami2019hyperbolic}} The Airport dataset encapsulates an international flight network with nodes signifying airports and edges representing flight routes. Node attributes encode airport-specific data, and classifications denote demographic information of the airport's country.
\paragraph{Cora\cite{sen2008collective}} This dataset comprises a scientific paper citation network characterized by nodes representing the papers, edges denoting citation linkages, and node attributes consisting of a bag-of-words representation. 
\paragraph{Cornell, Texas, and Wisconsin\cite{craven2000learning}} Deriving from the WebKB collection by Carnegie Mellon University, these subsets pertain to webpages from respective university computer science departments. Nodes embody webpages, edges correspond to hyperlinks, node attributes are derived from a bag-of-words model of the content, and node categories span five types: Student, Project, Course, Staff, and Faculty.
\paragraph{Chameleon and Squirrel\cite{rozemberczki2021multi}} Curated by Rozemberczki \etal, these datasets illustrate networks of Wikipedia pages on specific subjects. Nodes capture English Wikipedia articles linked by mutual hyperlinks. Node attributes reflect informational content and significant nouns, with labels classified by average monthly page traffic into five levels.
\paragraph{Actor\cite{tang2009social}} This dataset samples from a broader Films-Directors-Actors network, focusing exclusively on actors. Nodes indicate actors, edges represent co-mentions on Wikipedia, and attributes include key terms from the actors' Wikipedia articles. Node labels categorize actors into five groupings based on the lexical content of their associated Wikipedia pages.


\section{Sensitivity Analysis for $\beta$}

\begin{figure}[ht]
  \centering
  \begin{subfigure}{0.48\linewidth}
    \includegraphics[width=2.25in]{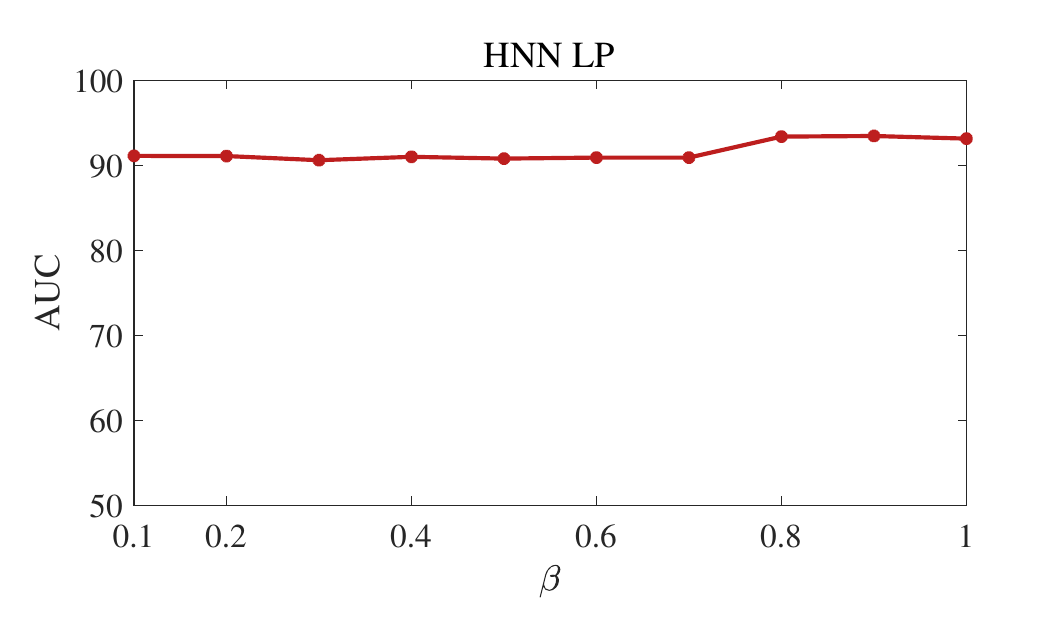}
    \caption{HNN-LP}
    \label{HNN_LP}
  \end{subfigure}
  \hfill
  \begin{subfigure}{0.48\linewidth}
    \includegraphics[width=2.25in]{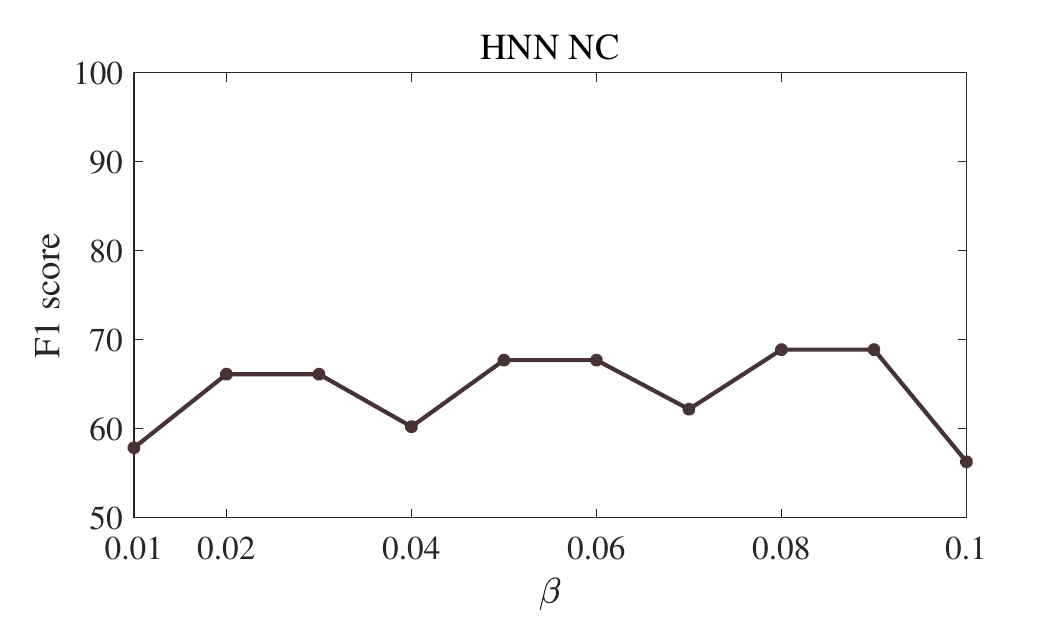}
    \caption{HNN-NC}
    \label{HNN_NC}
  \end{subfigure}
  \caption{Sensitivity analysis of HNN.}
  \label{HNN_sa}
\end{figure}

\begin{figure}[ht]
  \centering
  \begin{subfigure}{0.48\linewidth}
    \includegraphics[width=2.25in]{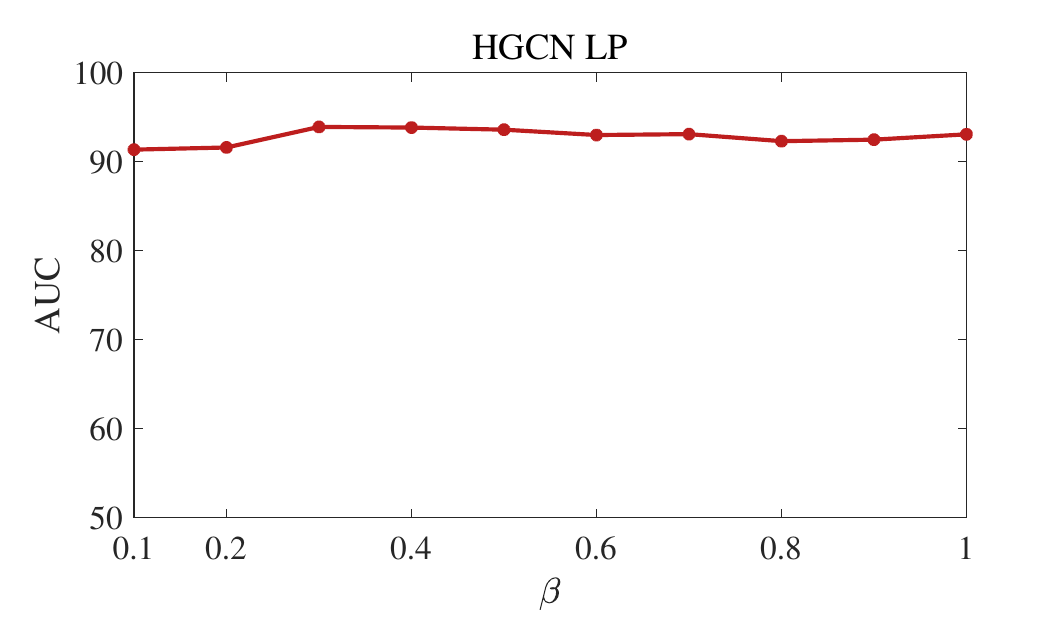}
    \caption{HCNN-LP}
    \label{HGCN_LP}
  \end{subfigure}
  \hfill
  \begin{subfigure}{0.48\linewidth}
    \includegraphics[width=2.25in]{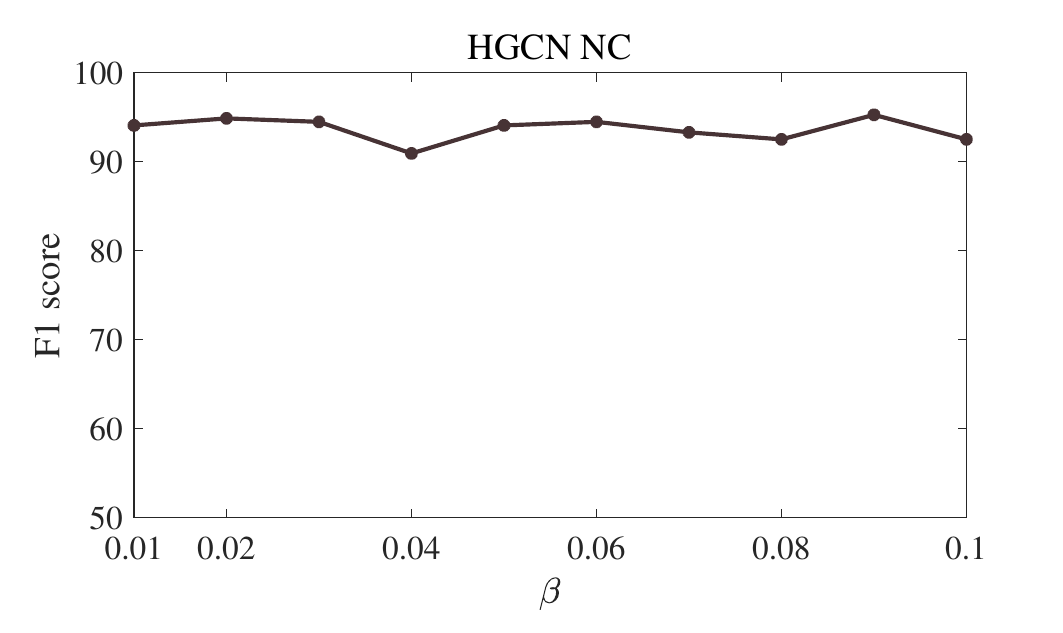}
    \caption{HCNN-NC}
    \label{HGCN_NC}
  \end{subfigure}
  \caption{Sensitivity analysis of HGCN.}
  \label{HGCN_sa}
\end{figure}

\begin{figure}[ht]
  \centering
  \begin{subfigure}{0.48\linewidth}
    \includegraphics[width=2.25in]{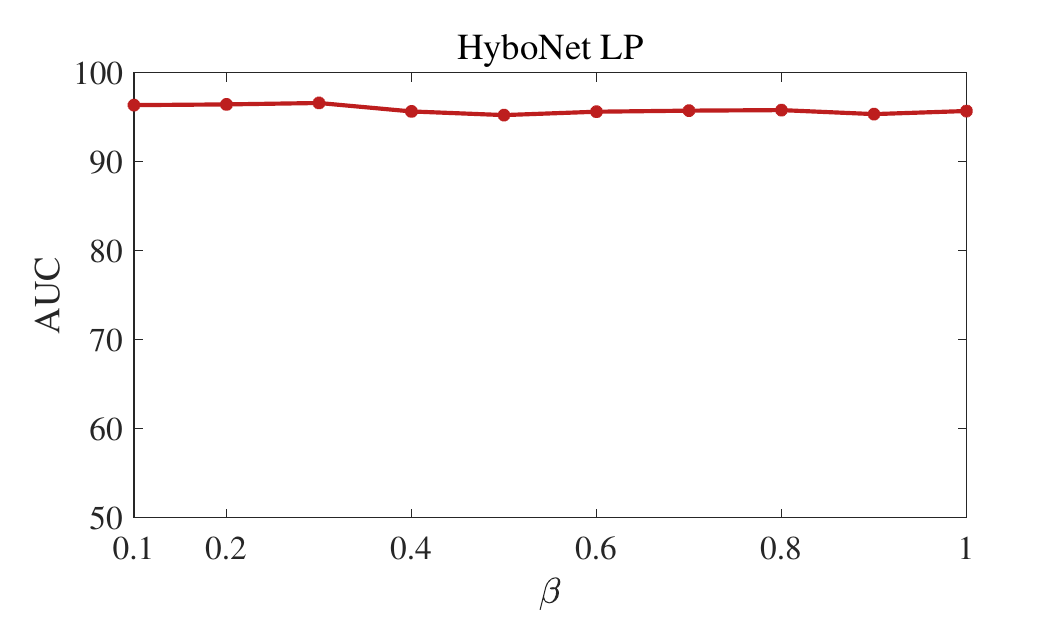}
    \caption{HyboNet-LP}
    \label{HyboNet_LP}
  \end{subfigure}
  \hfill
  \begin{subfigure}{0.48\linewidth}
    \includegraphics[width=2.25in]{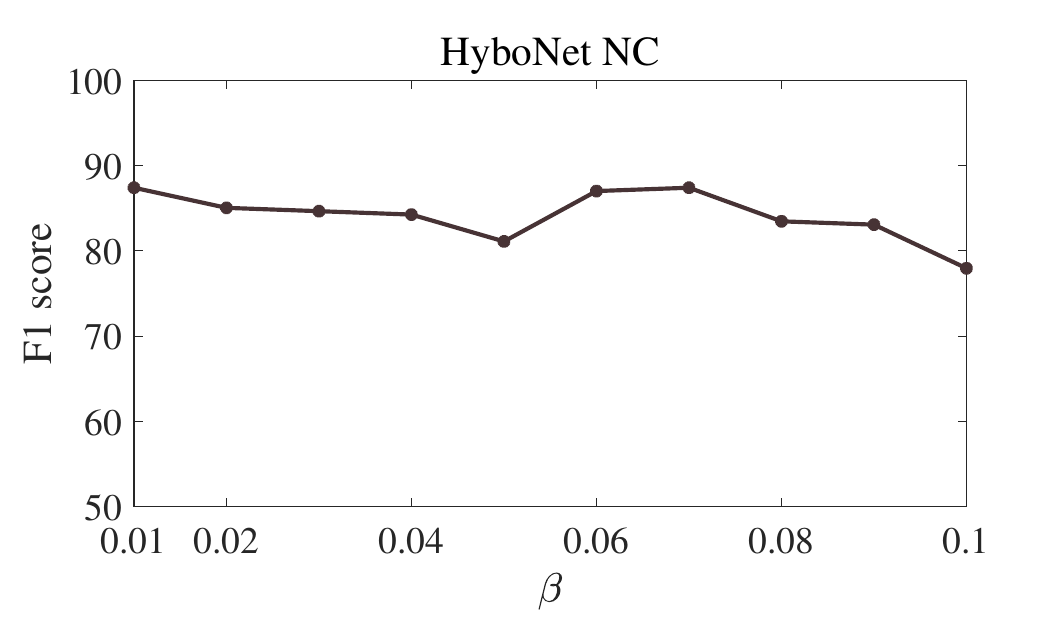}
    \caption{HyboNet-NC}
    \label{HyboNet_NC}
  \end{subfigure}
  \caption{Sensitivity analysis of HyboNet.}
  \label{HyboNet_sa}
\end{figure}

We conduct a sensitivity analysis for the hyperparameter $\beta$ on the Disease-LP and Disease-NC datasets. The effects of varying $\beta$ values on the performance are shown in Figs.~\ref{HNN_sa}--\ref{HyboNet_sa}.

The analysis reveals a relatively stable performance across different $\beta$ selections, indicating that the results are not highly sensitive to this hyperparameter. While certain $\beta$ values other than reported $\beta$ values occasionally yield enhanced performance, the validated choices still consistently outperform the baseline across tests. 

\section{Effectiveness of the GW regularization method}

\begin{figure}[ht]
    \centering
    \includegraphics[width=0.49\linewidth]{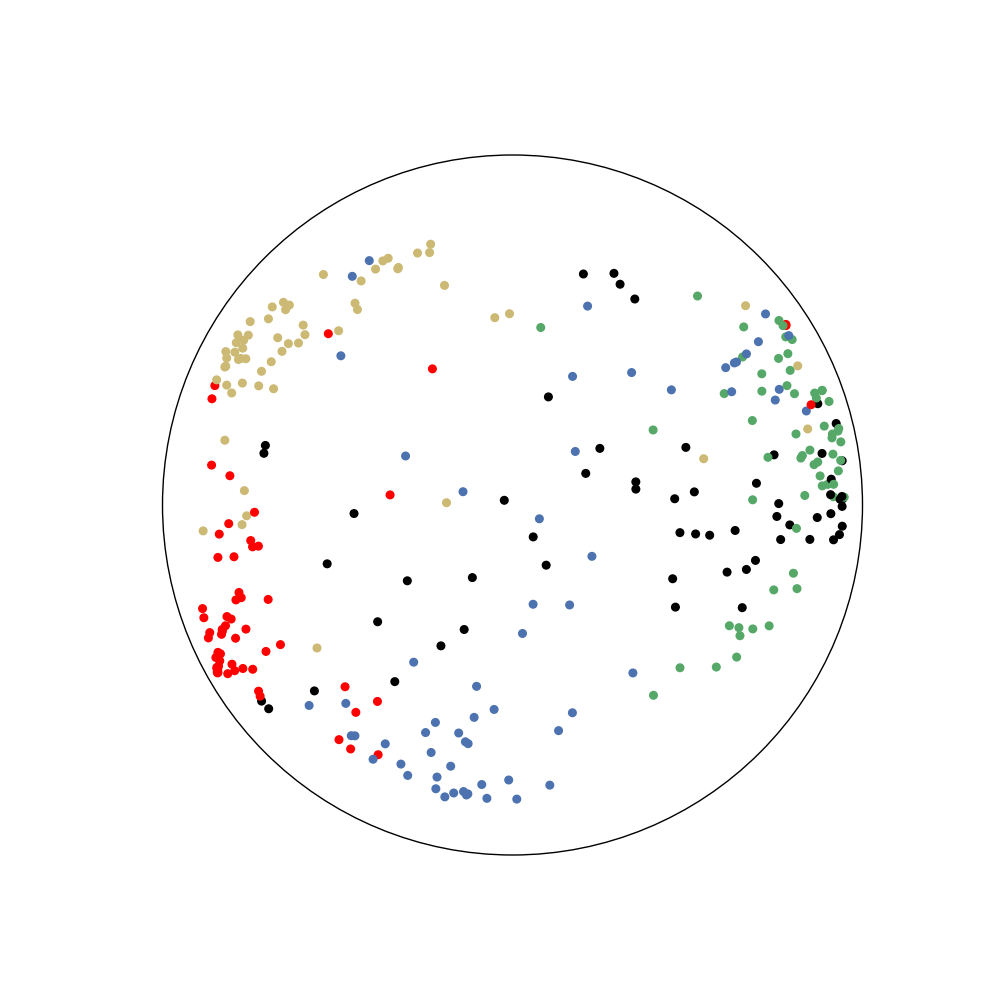}
    \includegraphics[width=0.49\linewidth]{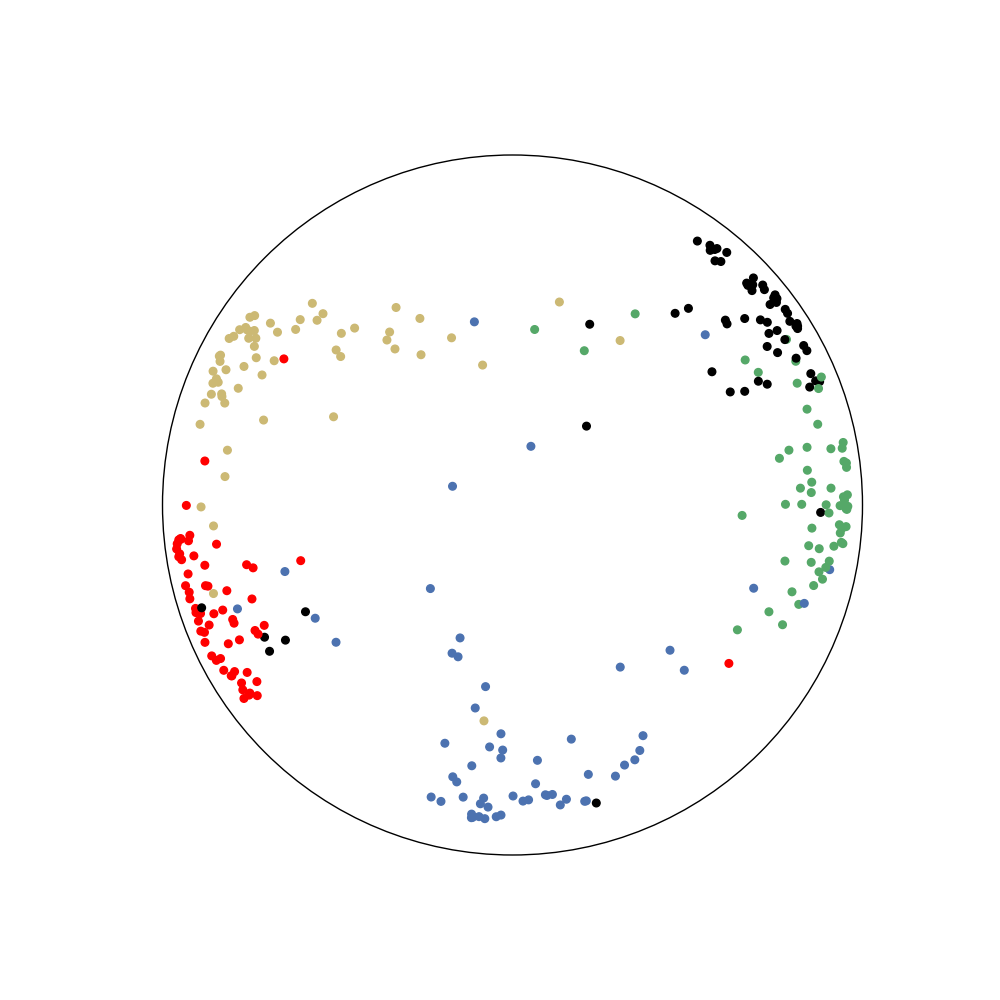}
    \caption{Output features of HNN layers for MiniImagenet (5-Shot-5-Way, also showing unused training samples). Different classes are shown in different colors. Left: No regularization 
    Right: GW regularization.} 
    \label{fig:visualization}
\end{figure}

We further validate the effectiveness of our method. First, Fig.~\ref{fig:visualization} visualizes the output of the HNN layers for MiniImagenet in 2D, where the dimension reduction is done by UMAP~\cite{mcinnes2018umap-software}. 
Clearly, with GW regularization, the features are better clustered and separated (the colors represent clusters).
Second, Table~\ref{tab:show_GW} shows the GW distances between the input and output features of the HNN layers, and implies that the GW distances are much smaller after regularization.

\begin{table}[ht]
\centering
\caption{GW distances between input and output of HNN layers}
\label{tab:show_GW}
\small
\begin{tabular}{cccc} 
\toprule
             & Dataset & No Reg & GW Reg     \\ 
\midrule
\multirow{3}{*}{\shortstack[c]{Image data \\ (5-Shot-5-Way \\ using Conv4)}} & {MiniImagenet} & {170.746}	&  {1.851}   \\ 
 & CUB          &  47833.570	&  1.575   \\ 
 & & & \\
\midrule
\multirow{9}{*}{\shortstack[c]{Graph data \\ (NC using HGCN)}} & Disease      & 8.646   & 5.604  \\ 
& Airport      & 0.136   & 0.040  \\ 
& Cora         & 4.059   & 0.292  \\ 
& Chameleon    & 3.879  & 0.316  \\ 
& Squirrel     & 1.211   & 0.100  \\ 
& Cornell      & 3.301   & 0.374  \\ 
& Texas        & 3.227   & 0.348  \\ 
& Wisconsin    & 2.958   & 1.828  \\ 
& Actor        & 0.315   & 0.090  \\
\bottomrule
\end{tabular}
\end{table}

\end{document}

%% file: math_commands.tex
\usepackage{amsmath,amsfonts,bm}
\usepackage{xcolor}

\def\eqref#1{(\ref{#1})}

\def\1{\bm{1}}

\def\rvx{{\mathbf{x}}}
\def\rvy{{\mathbf{y}}}
\def\rvz{{\mathbf{z}}}

\def\rmM{{\mathbf{M}}}

\def\mI{{\bm{I}}}

\DeclareMathAlphabet{\mathsfit}{\encodingdefault}{\sfdefault}{m}{sl}
\SetMathAlphabet{\mathsfit}{bold}{\encodingdefault}{\sfdefault}{bx}{n}

\def\gH{{\mathcal{H}}}

\def\sB{{\mathbb{B}}}

\def\sH{{\mathbb{H}}}

\def\sL{{\mathbb{L}}}

\def\sX{{\mathbb{X}}}

\newcommand{\R}{\mathbb{R}}

\newcommand{\abs}[1]{\left\lvert #1 \right\rvert}
\newcommand{\ip}[2]{\left\langle#1,#2\right\rangle}